\documentclass[10pt,twocolumn,letterpaper]{article}

\usepackage{cvpr}              
\usepackage{amsmath}
\usepackage{tabularx}
\usepackage{hhline}
\usepackage{float}
\usepackage{multirow}
\usepackage{booktabs}
\usepackage{balance} 
\usepackage{comment}

\usepackage{listings}
\usepackage{xcolor}
\usepackage{algorithm}
\usepackage{indentfirst}
\makeatletter
\renewcommand\@tocrmarg{2.55em plus1fil}
\makeatother
\definecolor{cvprblue}{rgb}{0.21,0.49,0.74}
\usepackage[pagebackref,breaklinks,colorlinks,allcolors=cvprblue]{hyperref}

\def\paperID{14924} 
\def\confName{CVPR}
\def\confYear{2025}

\title{Resilient Sensor Fusion under Adverse Sensor Failures  via  Multi-Modal Expert Fusion}

\makeatletter
\def\@fnsymbol#1{\ensuremath{\ifcase#1\or
  *\or \ddagger\or \dagger\or
  \mathsection\or \mathparagraph\or \|\or
  **\or \dagger\dagger\or \ddagger\ddagger
  \else\@ctrerr\fi}}
\makeatother

\author{
Konyul Park$^{1}$\thanks{Equal contributions} \quad
Yecheol Kim$^{2,3*}$\thanks{Work done during an internship at LG AI Research} \quad
Daehun Kim$^{2}$ \quad
Jun Won Choi$^{1}$\thanks{Corresponding author} \\
$^1$Seoul National University \quad $^2$Hanyang University \quad $^3$LG AI Research\\
{\tt\small kypark@spa.snu.ac.kr, \{yckim, dhkim\}@spa.hanyang.ac.kr, junwchoi@snu.ac.kr}
}

\begin{document}
\maketitle
\begin{abstract}

Modern autonomous driving perception systems utilize complementary multi-modal sensors, such as LiDAR and cameras. Although sensor fusion architectures enhance performance in challenging environments, they still suffer significant performance drops under severe sensor failures, such as LiDAR beam reduction, LiDAR drop, limited field of view, camera drop, and occlusion. This limitation stems from inter-modality dependencies in current sensor fusion frameworks. In this study, we introduce an efficient and robust LiDAR-camera 3D object detector, referred to as MoME, which can achieve robust performance through a mixture of experts approach. Our MoME fully decouples modality dependencies using three parallel expert decoders, which use camera features, LiDAR features, or a combination of both to decode object queries, respectively. We propose Multi-Expert Decoding (MED) framework, where each query is decoded selectively using one of three expert decoders.  MoME utilizes an Adaptive Query Router (AQR) to select the most appropriate expert decoder for each query based on the quality of camera and LiDAR features. This ensures that each query is processed by the best-suited expert, resulting in robust performance across diverse sensor failure scenarios. We evaluated the performance of MoME on the nuScenes-R benchmark. Our MoME achieved state-of-the-art performance in extreme weather and sensor failure conditions, significantly outperforming the existing models across various sensor failure scenarios.
\end{abstract}    
\begin{figure}[t]
   \centering
   \includegraphics[width=1.0\columnwidth]{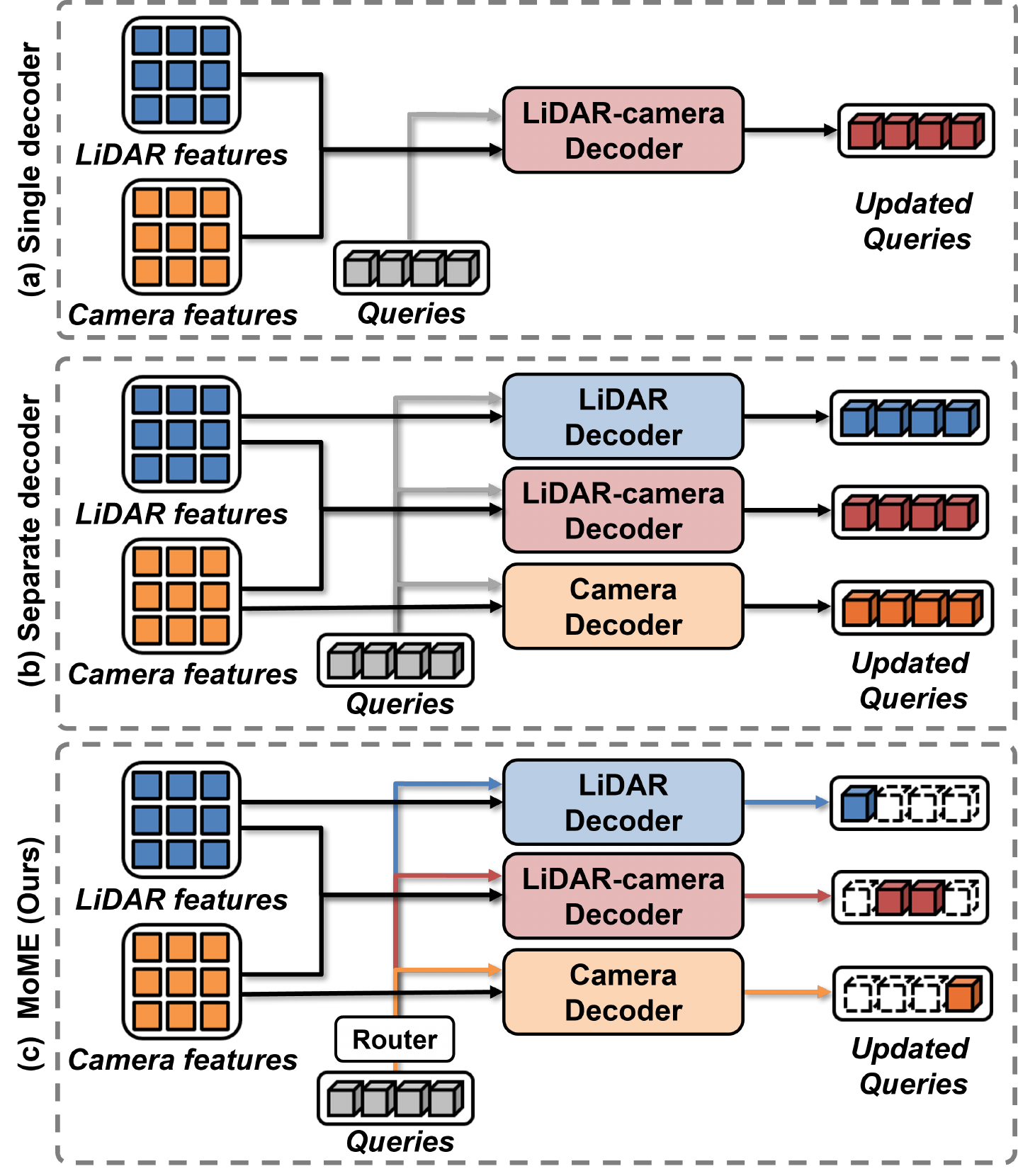}
   \caption{\textbf{Comparisons of several LiDAR-camera fusion methods.} (a) Single decoder, (b) Separate decoder, and (c) Proposed MoME. MoME utilizes three parallel decoders, each processing camera features, LiDAR features, or a combination of both, to decouple dependencies between the modalities. It adaptively routes each query to the most suitable decoder based on the quality of the camera and LiDAR features. }
   \label{intro}
\end{figure}
\begin{figure}[t]
   \centering
   \includegraphics[width=1.0\columnwidth]{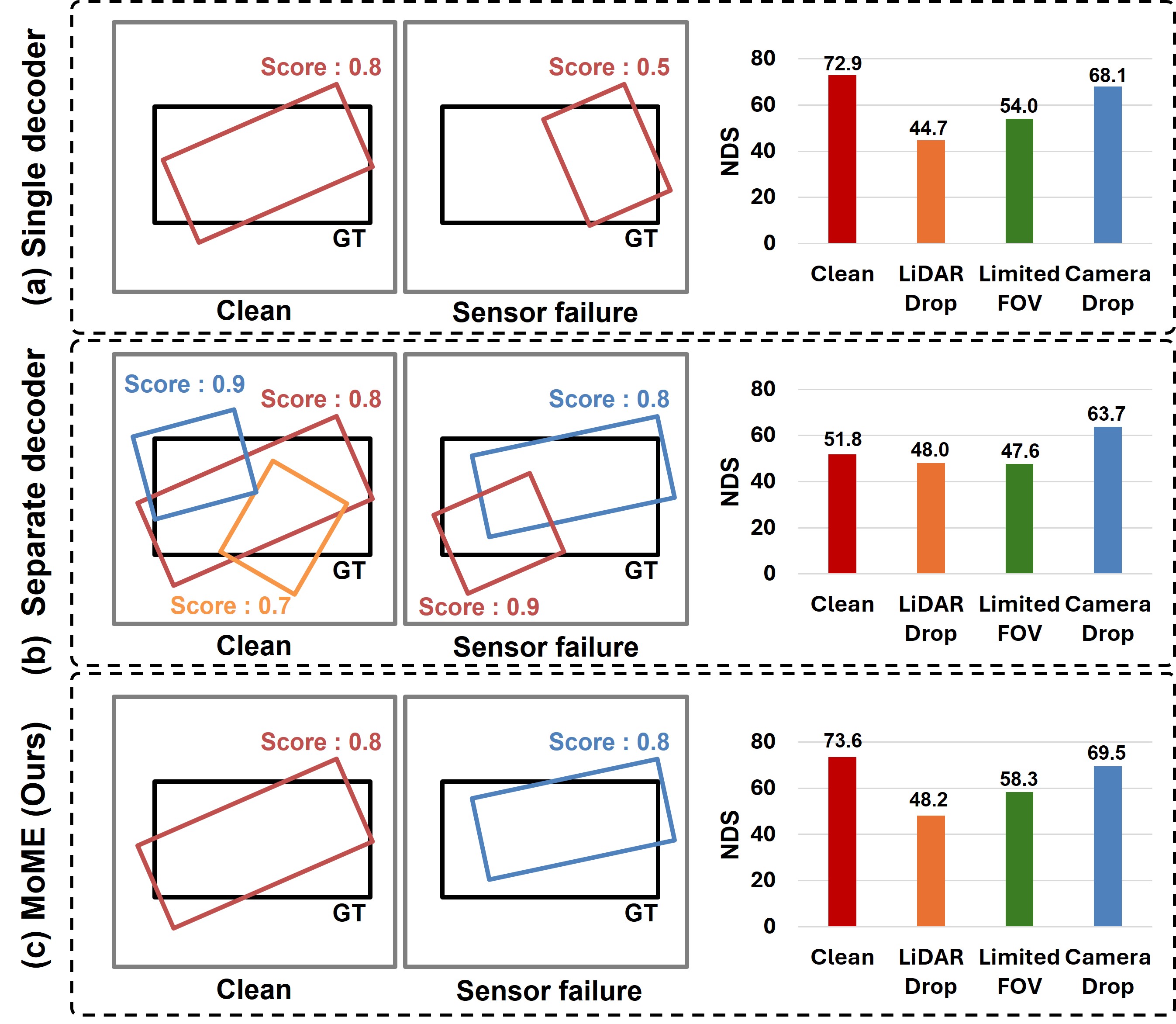}
   \caption{\textbf{Comparison of predictions and performance of different fusion methods.} Left: Examples of detection results for a single object. For visualization, red boxes indicate predictions from the LiDAR-camera decoder, blue boxes from the LiDAR decoder, and orange boxes from the camera decoder, with ground truth (GT) shown in black. Right: NDS performance across various scenarios—Clean, LiDAR drop, Limited FOV, and Camera drop. (a) Under sensor failure, predictions are prone to error due to the corrupted modality. (b) Due to uncalibrated confidence score scales, lower-quality predictions may have higher confidence scores. (c) The proposed method selects the optimal decoder for decoding each object query.      }
   \label{intro2}
   \vspace{-0.3cm}
\end{figure}
\section{Introduction}
\label{sec:intro} 

Accurate 3D perception is crucial for the safe and reliable operation of autonomous vehicles \cite{pointpillar, second,transfusion}. Multi-modal perception systems enable more reliable understanding of environments by leveraging the complementary characteristics of different sensors \cite{bevfusion_ADLAB, 3d-cvf, pointpainting,cmt}. Each sensor modality brings distinct benefits: cameras capture rich semantic information, LiDAR provides precise 3D geometric details, and radar ensures reliable long-range detection even under adverse weather conditions. Consequently, sensor fusion plays a pivotal role in enhancing perception robustness.


Recent advances in multi-modal perception have mainly focused on intermediate-level fusion strategies, which combine intermediate features extracted from multiple modalities \cite{deepinteraction, sparsefusion, bevfusion_MIT, bevfusion_ADLAB, 3d-cvf, cmt}. This fusion strategy can improve the quality of intermediate features, contributing to robust performance across various scenarios. However, their performance can degrade significantly when one sensor is affected by severe noise, occlusion, or failure. This limitation arises from inter-modal dependencies, where contamination in one sensor can impair the overall quality of multi-modal perception. Such dependencies are challenging to avoid in current fusion models, as utilization of high-level interactions between sensors is essential for effective information fusion.


As autonomous vehicle safety becomes increasingly critical, ensuring robustness under sensor failure conditions has emerged as a key challenge in the field. Sensor failures encompass a range of real-world issues, including complete sensor malfunctions, partial impairments (such as a reduced field of view or decreased resolution), and adverse conditions (such as occlusion or contamination). These challenges can significantly degrade perception performance, even when sensor fusion methods are employed.

Recent studies \cite{metabev, unibev, robbev, cmt} aimed to facilitate robustness to such sensor failures for 3D perception. MetaBEV \cite{metabev} proposed a robust sensor fusion technique that applies successive cross-modal attention using multi-modal features. RobBEV \cite{robbev} mitigated inter-modality interference by employing mutual deformable attention, where each modality queries features from the other and vice versa. UniBEV \cite{unibev} aligned camera and LiDAR features adaptively through channel-normalized weights. CMT \cite{cmt} employed position-encoded cross-attention for feature fusion, combined with masked-modal training. 
While these methods have effectively mitigated the impact of sensor failures, we argue that they have not fully resolved the issue, as the features from the two modalities remain coupled, leading to inevitable performance drops (see Fig.~\ref{intro2} (a)). Consequently, the performance of sensor fusion may often fall short of what could be achieved with a single, fully functioning sensor.

This paper addresses the aforementioned challenges in LiDAR-camera fusion, a commonly used sensor suite in autonomous driving and robotics. Inspired by the Mixture of Experts (MoE) framework \cite{moe}, which has been widely adopted in natural language processing to enhance model performance and efficiency, we introduce the Multi-Expert Decoding (MED) framework. In MoME, three parallel Transformer decoders, called  expert decoders independently utilize camera features, LiDAR features, and their combination to decode shared object queries (see Fig.~\ref{intro} (b) and (c)). This approach decouples modality dependencies, protecting the detector from the impacts of sensor failures.

A straightforward way to utilize these expert decoders is to have them process the queries in parallel \cite{meformer} and select the output based on a confidence score, as shown in Fig.~\ref{intro} (b). However, this approach poses two major challenges. First, as illustrated in Fig.~\ref{intro2}, the decoders operate independently, leading to uncalibrated confidence scores, which complicates fair comparison and results in suboptimal performance. Second, the decoding complexity is tripled, imposing a  computational burden on 3D perception. Our study addresses these issues to present an efficient, robust sensor fusion architecture.

In this paper, we propose a novel LiDAR-camera fusion model for 3D object detection, named {\it Mixture of Multi-modal Expert} (MoME), designed to be resilient under adverse sensor failure conditions. {\it MoME} builds upon DETR \cite{detr}, utilizing a Transformer to progressively decode a fixed set of object queries into detection results across multiple layers. To enhance robustness, {\it MoME} incorporates the MED framework into DETR, where three parallel decoders are used as expert modules, as illustrated in Fig.~\ref{intro} (c).

To enable an efficient decoding process, MoME selectively assigns each object query to one of three expert decoders. This selection is managed by the Adaptive Query Router (AQR), which dynamically determines the most suitable expert based on the quality of LiDAR and camera features.
AQR extracts positional information from each query and aggregates multi-modal features within a local region identified by the Local Attention Mask. It then applies cross-attention with these local features to compute modal selection probabilities, ensuring that each query is processed by the most appropriate expert decoder.
By training with synthetic sensor drop augmentation, our MED framework learns to select the expert decoder that optimizes performance. Additionally, by assigning each query to only one of the three decoders, our approach maintains a decoding complexity comparable to that of a single decoder, as shown in Fig.~\ref{intro} (a).

We evaluated the performance of {\it MoME} on widely used nuScenes dataset \cite{nuscenes}. We followed nuScenes-R benchmark \cite{robustbench}, which includes six critical sensor failure scenarios. {\it MoME} achieved state-of-the-art (SOTA) performance across most sensor failure scenarios.
Notably, compared to the latest SOTA method, CMT \cite{cmt}, {\it MoME} demonstrates significant performance gains, with \textbf{+4.2}\% mAP in LiDAR-drop scenarios, \textbf{+1.9}\% mAP in camera-drop scenarios, and \textbf{+6.7}\% mAP in limited FOV scenarios. Additionally, {\it MoME}’s superior performance under extreme weather conditions underscores its potential to enhance robustness in adverse real-world environments.

The key contributions of our study are summarized as follows:
\begin{enumerate}
   \item We propose a new LiDAR-camera fusion architecture resilient to sensor failures. Our {\it MoME} method employs three parallel expert decoders that separately utilize camera features, LiDAR features, and a combination of both. This approach effectively decouples inter-modality dependence.
   \item We propose a MED framework in which one of three decoders is selected to decode each query. To this goal, we devise an AQR model that produces the probability of selecting the expert decoder based on the quality of modality-specific features. 
   \item The multi-expert decoder structure in MoME is straightforward to implement and can be readily integrated into any Transformer-based feature fusion architecture. Furthermore, as each query is assigned to only one expert decoder, the computational overhead of MoME remains minimal. 
    \item MoME achieved SOTA performance among existing camera-LiDAR fusion methods under various sensor failure conditions and extreme weather scenarios.
   \item The code will be publicly available. 
\end{enumerate}
\begin{figure*}[h]
    \centering
    {\includegraphics[width=0.95 \textwidth]{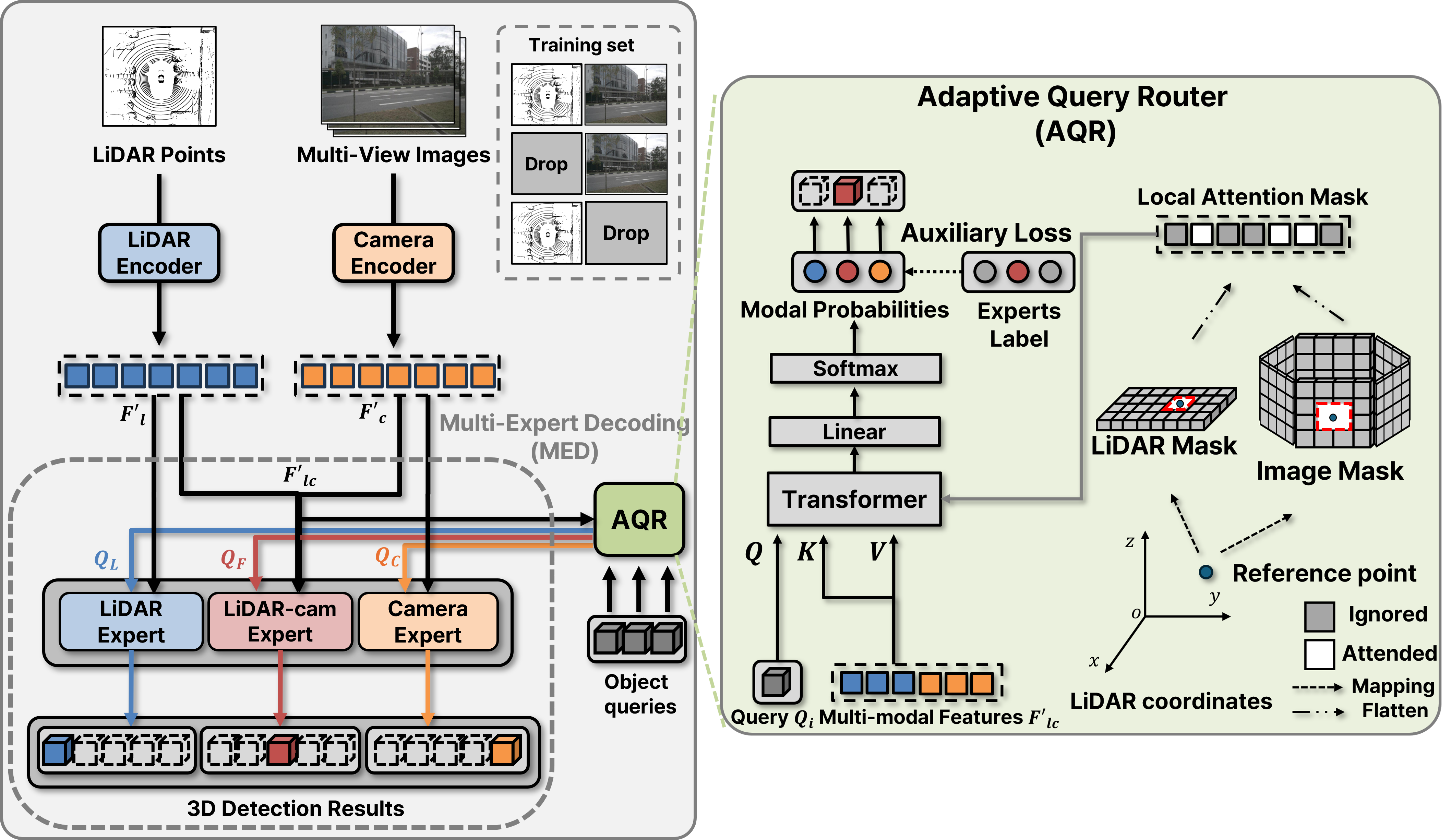}}
    \hspace{5mm}
    \caption {\textbf{Overall structure of MoME.} MoME utilizes three expert decoders, each specialized for processing LiDAR, camera, or LiDAR-camera features. The AQR dynamically assigns each object query to the most suitable expert decoder in an adaptive manner. The operation of AQR relies on local features filtered through the Local Attention Mask.}
    \label{overall}
\end{figure*}
\section{Related Works}
\label{sec:formatting}



\subsection{Sensor Failure}


Recent benchmarks \cite{robustbench, robo3d, benchmarking_cc} have introduced comprehensive evaluation protocols for sensor failures and degradation, addressing the limitations of existing autonomous driving datasets that primarily focus on ideal driving conditions. Several approaches have been proposed to maintain robust perception under sensor failures. MetaBEV \cite{metabev} employs modality-arbitrary BEV-evolving decoder with cross-modal attention, while UniBEV \cite{unibev} proposes channel-wise normalization for BEV feature alignment. CMT \cite{cmt} and MetaBEV \cite{metabev} leverage sensor dropout as data augmentation to improve robustness against sensor failures. However, the joint multi-modal learning scheme in existing methods creates inter-modal dependencies, leading to cascading performance degradation under sensor failures. In this work, we propose MED framework that processes each modality independently, enabling selective handling of sensor corruption while preserving cross-modal complementarity.

\subsection{Mixture of Experts}

MoE \cite{moe} enables linear scaling of model capacity with minimal computational complexity by dynamically selecting top-K experts from a pool of independent networks. Each expert specializes in distinct patterns, facilitating efficient decomposition of complex tasks into specialized sub-networks. However, this multi-expert routing mechanism often leads to training instability and convergence issues. Switch Transformer \cite{switchtransformer} addressed these limitations by implementing single-expert routing, significantly improving training efficiency and scalability. The MoE approach has recently received attention in computer vision applications. V-MoE \cite{v-moe} extended the concept to Vision Transformers \cite{vit}. MetaBEV \cite{metabev} leveraged MoE architecture to mitigate gradient conflicts in multi-task learning scenarios. 

In this work, we extends the MoE paradigm to multi-modal perception tasks, by leveraging its capability  to handle heterogeneous inputs through specialized expert networks. This approach naturally aligns with sensor fusion scenarios where input reliability varies dynamically.

\section{Proposed Method}
\subsection{Overview}
The overall architecture of MoME is depicted in Fig.~\ref{overall}. We employ VoxelNet \cite{voxelnet} as a backbone  to encode LiDAR points $P_l \in \mathbb{R}^{M \times 3}$. The network first discretizes $P_l$ into 3D voxels, then hierarchically extracts voxel features through a series of 3D sparse convolutions. The voxel features are then projected onto a BEV  and processed through a series of 2D convolutional layers to generate BEV features $F_l \in \mathbb{R}^{H_l \times W_l \times D}$, where ($H_l$, $W_l$) is the size of BEV features and $D$ is the channel dimension. 
We utilize VoVNet \cite{lee2020centermask} to encode multi-view camera images separately. It generates 2D multi-view camera features  $F_c \in \mathbb{R}^{V \times H_c \times W_c \times D}$, where $V$ is the number of camera views, ($H_c$, $W_c$) is the size of 2D camera features, and $D$ is the channel dimension. 
These complementary features $F_l$ and $F_c$ are flattened into $F'_l \in \mathbb{R}^{(H_l W_l) \times D}$ and $F'_c \in \mathbb{R}^{(V H_c W_c) \times D}$, respectively. These two flattened features are then concatenated to obtain the fused LiDAR-camera features $F'_{lc} \in \mathbb{R}^{(H_l W_l + V H_c W_c) \times D}$.

To produce detection results, MoME decodes object queries \( Q \in \mathbb{R}^{N \times D} \) using \( F'_l \), \( F'_c \), and \( F'_{lc} \) as keys and values. For robust decoding, we employ three expert decoders: the \textit{LiDAR decoder}, the \textit{camera decoder}, and the \textit{LiDAR-camera decoder}. Specifically, the LiDAR decoder performs cross-attention using \( F'_l \), the camera decoder uses \( F'_c \), and the LiDAR-camera decoder uses \( F'_{lc} \).
The MED framework assigns each of the \( N \) queries to one of the three expert decoders which is selected by the AQR module. AQR selects the appropriate decoder for each query based on local features pooled from $F'_{lc}$ centered at the position indicated by each query.
As a result, the MED framework partitions the queries into three groups, which are then processed by the LiDAR decoder, camera decoder, and LiDAR-camera decoder, respectively. Finally, the decoded queries are used to generate 3D bounding boxes and class scores.


\subsection{Multi-Expert Decoding}

The structure of the MED framework is illustrated in Fig.~\ref{overall}. 
The MED framework consists of two main components: (1) MED that employs three expert decoders capable of independently processing queries using LiDAR features, camera features, or a combination of both, and (2) AQR that dynamically assigns each query to the most suitable expert decoder in an adaptive manner. Each expert decoder adopts the decoder structure from DETR. 

Let $q_i \in \mathbb{R}^{1 \times D}$ represent the $i$-th object query within the set $Q$ of object queries. The AQR processes both $q_i$ and the multi-modal features $F'_{lc}$ to produce the modal selection probabilities $\mathbf{p}_i = [p_{i,l}, p_{i,c}, p_{i,lc}]$. The expert decoder responsible for decoding the $i$-th query is determined based on the probability with the highest value. Consequently, the object queries are grouped into $Q_c$, $Q_l$, and $Q_{lc}$, where 
\begin{align} \label{eqn:argmax}
Q_{mod} &= \begin{bmatrix} q_i  \in Q \mid \arg\max(\mathbf{p}_i) = mod \end{bmatrix},
\end{align}
and $mod \in \{l, c, lc\}$. 
These query groups are decoded in parallel using the corresponding expert decoders, resulting in the updated queries $Q'_c$, $Q'_l$, and $Q'_{lc}$. Finally, a Feed-Forward Network (FFN) is applied to these updated queries to predict the final detection results.

\subsection{Adaptive Query Router (AQR)}

AQR employs locality-aware adaptive routing to determine the most appropriate expert decoder for each query. This process begins by calculating a Local Attention Mask that identifies the local features in the LiDAR BEV and camera domains associated with the $i$-th query.

The construction of the Local Attention Mask is depicted in Fig.~\ref{overall}. The mask has the same dimensions as $F'_{lc}$ and the elements have binary values. Let the $i$-th object query $q_i$ be associated with a reference point $p_{i,ref} = (x, y, z)^T$ in 3D coordinates. This reference point is projected onto the LiDAR BEV domain resulting in $p^{(l)}_{i,ref} = (x_l, y_l)^T$. Similarly, it is also projected onto one of the multi-view camera domains producing $p^{(c)}_{i,ref} = (x_c, y_c)^T$. All elements of the Local Attention Mask are set to zero while those within an $l_l \times l_l$ window centered at $p^{(l)}_{i,ref}$ in the LiDAR BEV domain and within an $l_c \times l_c$ window centered at $p^{(c)}_{i,ref}$ in the camera view are set to one.

Finally, AQR applies masked cross-attention, where the binary Local Attention Mask serves as the mask, the multi-modal features $F'_{lc}$ are used as the key and value, and the set of queries $Q$ is used as the query. A softmax function is applied to the output of each decoded query to compute modal selection probabilities, which are then used to assign the query to the appropriate expert decoder.

\subsection{Training Details}
\label{sec:training_details}
To train MoME, it is crucial to guide the model to utilize the expert decoder unaffected by sensor failures. To achieve this, we randomly drop either the camera or LiDAR input during training, with each sensor having a $1/3$ probability of being dropped, and no sensor dropped for the remaining $1/3$ \cite{cmt}. Although we simulate complete sensor failure during training, our approach is also effective in partial sensor failure scenarios, as the routing operation of AQR relies on local features identified by the Local Attention Mask. Consequently, the appropriate expert decoder is selected based on the local features pointed by each query. Furthermore, our experiments in the next section demonstrate that our method, trained solely under these synthetic drop conditions, generalizes well to other types of sensor failures. Notably, our method is data-scalable, as it does not require generating various types of sensor failures for training data.     

The training process consists of three stages. In the first stage, all object queries in $Q$ are processed in parallel by each expert decoder without applying sensor drop data augmentation. The decoded queries from each expert decoder are matched with GT through bipartite matching, and all expert decoders are simultaneously trained using the loss function
\begin{align}
   \mathcal{L}_{1st} = \mathcal{L}_l + \mathcal{L}_c + \mathcal{L}_{lc},
\end{align}
where $\mathcal{L}_l$, $\mathcal{L}_c$, and $\mathcal{L}_{lc}$ represent the detection losses associated with LiDAR, camera, and LiDAR-camera decoders, respectively. We employ the focal loss \cite{focal} for classification and the L1 loss for box regression.

In the second stage, we freeze the pre-trained expert decoders and focus on training the AQR module. During this stage, we simulate sensor failure scenarios by randomly dropping LiDAR or camera sensors. Supervisory labels are generated as $\mathbf{y}_i = [0, 1, 0]$ when LiDAR is dropped, $\mathbf{y}_i = [1, 0, 0]$ when the camera is dropped, and $\mathbf{y}_i = [0, 0, 1]$ when neither sensor is dropped.
 We train AQR using 
\begin{align}
\mathcal{L}_{2nd} =  \sum_{i=1}^N \text{CE}(\mathbf{y}_i, \mathbf{p}_i),
\end{align}
where CE denotes the cross entropy loss between the modal selection probabilities $\mathbf{p}_i$ and the experts label $\mathbf{y}_i$. 

\section{Experiments}
\subsection{Experimental Setup}

\begin{table*}[!h]
    \centering
    \small
    \renewcommand{\arraystretch}{1.1}
    \resizebox{\textwidth}{!}{
     \begin{tabular}{@{}c|cc|cc|cc|cc|cc|cc|cc|cc@{}}
       \toprule
       \multicolumn{1}{c|}{} & \multicolumn{2}{c|}{}& \multicolumn{2}{c|}{}&\multicolumn{8}{c|}{LiDAR failures} & \multicolumn{4}{c}{Camera failures}\\
       \cline{6-13} \cline{14-17}
       \multirow{2}[-1]{*}{\normalsize Method} & \multicolumn{2}{c|}{Clean} & \multicolumn{2}{c|}{Perf. Ratio ($R$)} & \multicolumn{2}{c|}{Beam Reduction} & \multicolumn{2}{c|}{LiDAR Drop} & \multicolumn{2}{c|}{Limited FOV} & \multicolumn{2}{c|}{Object Failure} & \multicolumn{2}{c|}{View Drop} & \multicolumn{2}{c}{Occlusion} \\ 
        
        \multicolumn{1}{c|}{} & \multicolumn{2}{c|}{}& \multicolumn{2}{c|}{} & \multicolumn{2}{c|}{\textit{4 beams}} & \multicolumn{2}{c|}{\textit{all }} & \multicolumn{2}{c|}{\textit{[-60, 60]}} & \multicolumn{2}{c|}{\textit{rate = 0.5}} & \multicolumn{2}{c|}{\textit{6 drops}}& \multicolumn{2}{c}{\textit{w obstacle}} \\
        \cline{2-17}
        & mAP & NDS & mAP & NDS& mAP & NDS & mAP & NDS & mAP & NDS & mAP & NDS& mAP & NDS & mAP & NDS \\ 
        \hline
        DETR3D \cite{detr3d} &  34.9 & 43.4 & 20.5 &33.4  & - & - & -& - & - & - & - & - & 0.0 & 0.0 & 14.3 & 29.0  \\ 
        CenterPoint$^{\dagger}$ \cite{centerpoint} & 56.2 & 64.7 & 28.9 & 51.9& 20.9 & 42.9  & 0.0& 0.0 & 15.7 & 43.0 & 28.4 & 48.5 & - & - & - & -  \\ 
        UniBEV$^{\dagger}$ \cite{unibev} & 64.2 & 68.5 & 75.8 & 83.7 & 45.1 & 56.1 & 35.0 & 42.2 & 35.9 & 50.1 & 57.1 & 64.0 & 58.2 & 65.3 & 60.5 & 66.4\\
        TransFusion \cite{transfusion} & 66.9 & 70.9 & 54.4 & 66.8& - & -  & 0.0 & 0.0 & 20.3 & 45.8 & 34.6 & 53.6 & 61.6 & 67.4 & 65.5 & 70.0\\ 
        BEVFusion$^{\dagger}$ \cite{bevfusion_ADLAB} & 67.9 & 71.0& 55.2 & 67.8 & 43.0 & 55.4  & 0.0 & 0.0 & 21.1 & 45.6 & 39.2 & 54.6 & 56.2 & 63.5 & 65.3 & 69.6\\
        MetaBEV \cite{metabev} & 68.0 & 71.5& 75.4 & 82.5  & - & 57.7 & 39.0 & 42.6 & - & 47.0 & - & 67.6 & 63.6 & 69.2 & - & 70.0 \\
        BEVFusion$^{\dagger}$ \cite{bevfusion_MIT} & 68.5 & 71.4 & 62.1& 72.5 &46.4& 57.3 & 0.2 & 4.0 & 22.0 & 45.8 & 63.2 & 68.0 & 58.8 & 66.1 & 64.8 & 69.4 \\
        DeepInteraction$^{\dagger}$ \cite{deepinteraction} & 69.9 & 72.6 & 58.9 & 64.9 &46.5& 54.9 & 0.0 & 0.0 & 21.2 & 42.7 & 60.8 &63.7  & 53.9 & 55.1 & 64.5 & 66.2 \\
        
        CMT$^{\dagger}$ \cite{cmt} & 70.3 & 72.9& 78.4 & 84.4  & 54.9 & 62.2 & 38.3 & 44.7 & 43.9 & 54.0 & 66.7 & 70.4 & 61.7 & 68.1 & 65.0 & 69.8 \\
        UniTR$^{\dagger}$ \cite{unitr} & 70.5 & 73.3 &54.8&  68.5& 47.7 & 59.6 & 0.1 & 1.5 & 22.4 & 48.1 & 38.3 & 55.6 & 60.4 & 67.4& 62.7 & 68.9 \\
        SparseFusion$^{\dagger}$ \cite{sparsefusion} & 71.0 & 73.1& 62.4 & 71.3  &49.8& 59.6 & 0.0 & 0.0 & 24.0 & 45.6 & 65.2 & 69.5 & 59.7 & 67.2 & \textbf{67.2} & \textbf{71.0} \\
        \textbf{MoME (ours)} &  \textbf{71.2} & \textbf{73.6}& \textbf{80.5} & \textbf{86.2}  & \textbf{55.0} & \textbf{63.0} & \textbf{42.5} & \textbf{48.2} & \textbf{50.6} & \textbf{58.3} & \textbf{67.0} & \textbf{71.0} & \textbf{63.6} & \textbf{69.5} & 65.6 & 70.5 \\ 
        \bottomrule
    \end{tabular}
    }
    \caption{\textbf{Comparison under various sensor failure scenarios on nuScenes-R dataset.}  \textit{Italic} denotes the degree of sensor failure. Note that DETR3D uses only cameras, while CenterPoint uses only LiDAR. '$\dagger$' indicates reproduced results using their open-source code. MoME achieves state-of-the-art performance across most tasks, significantly surpassing prior methods in the relative performance ratio $R$.}

\label{table_2}
\end{table*}

\noindent\textbf{Datasets and metrics.} 
We evaluate the performance of MoME using the nuScenes dataset  \cite{nuscenes} and its variants, nuScenes-R \cite{robustbench} and nuScenes-C \cite{benchmarking_cc}. 
nuScenes \cite{nuscenes} is a driving dataset collected using a 32-beam LiDAR scanner operating at 20 Hz and six cameras providing 360° coverage. In total, it consists of 1.4 million camera images, 390 thousand LiDAR sweeps, and 1.4 million annotated 3D bounding boxes across 40 thousand keyframes, with annotations at 2Hz. The dataset includes 700 training scenes, 150 validation scenes, and 150 test scenes.

nuScenes-R \cite{robustbench} is a variant of the nuScenes dataset, created by simulating sensor failures for LiDAR and camera inputs. In our experiments, we consider five types of sensor failure scenarios:
\begin{itemize}
\item \textbf{LiDAR Drop}: All LiDAR point clouds are removed. 
\item \textbf{Limited FOV}: The scanning angle is restricted to a range of $-60^\circ$ to $60^\circ$.
\item \textbf{Object Failure}: LiDAR points within each bounding box are removed with a 50\% probability.
\item \textbf{Camera View Drop}: All multi-view camera images are set to zero.
\item \textbf{Occlusion}: Predefined mud masks are applied to the images.
\end{itemize}
These scenarios enable us to evaluate models under different sensor failure conditions.

nuScenes-C \cite{benchmarking_cc} systematically generates 27 types of common corruptions in 3D object detection for both LiDAR and
camera sensors. The corruptions are grouped into weather, sensor, motion, object, and alignment levels, covering the majority of real-world corruption cases. Our experiments consider extreme weather conditions including dense fog, snow, and sunlight glare.

Our model was trained on the nuScenes \cite{nuscenes} training set with simulated camera and LiDAR drop scenarios as described in Section \ref{sec:training_details}. It was then evaluated on the nuScenes validation set, as well as on nuScenes-R \cite{robustbench} and nuScenes-C \cite{benchmarking_cc}.

We adopted two widely used performance metrics for nuScenes: mean Average Precision (mAP) and the nuScenes Detection Score (NDS). mAP was computed from a bird's-eye view over 10 object classes, with distance thresholds of \{0.5$m$, 1$m$, 2$m$, 4$m$\}. NDS was obtained by combining mAP with mean errors in translation, scale, orientation, velocity, and attributes. Additionally, we evaluated the robustness using a relative performance ratio $R = {mP_R}/{P_C}$, where $P_C$ denotes the clean performance and $mP_R$ represents the average performance across adverse conditions. The $mP_R$ is given by
\begin{equation}
mP_R = \frac{1}{N} \sum_{i=1}^{N} P_{R}^{i}.
\end{equation}
where $N$ represents the total number of adverse cases considered and $P_{R}^{i}$ denotes the performance evaluated under the $i$-th adverse condition among the $N$ cases.

\noindent\textbf{Implementation details.} We adopt VoVNet \cite{lee2020centermask} and VoxelNet \cite{voxelnet} as the image and LiDAR backbones, respectively. 
In our architecture, the expert decoders utilize a 6-layer Transformer decoder with shared weights, while the AQR module consists of a single cross-attention layer. The attention mask sizes $l_l$ and $l_c$ were empirically set to 5 and 15, respectively. Our training procedure consists of two stages: We first trained the modality backbones and expert decoders for 20 epochs. Subsequently, we froze these networks and trained the AQR module for additional two epochs. While training the model, we employed Class-balanced Grouping and Sampling (CBGS)  \cite{cbgs} and data augmentation, as suggested in \cite{cmt}. The CBGS strategy was applied during the first stage only. For data augmentation, we considered random flipping along X and Y axes, rotation, scaling, translation, and ground-truth sampling.

\subsection{Performance Comparison}
Table \ref{table_2} presents the performance of MoME on both the nuScenes validation set and the nuScenes-R dataset. Under sensor failure conditions, MoME significantly outperforms the latest robust sensor fusion methods designed to handle such failures, including MetaBEV \cite{metabev}, CMT \cite{cmt}, and UniTR \cite{unitr}. MoME achieves substantial improvements in the relative performance ratio $R$ with gains of +2.1\% and +1.8\% over the previous SOTA method, CMT \cite{cmt}. Notably, MoME demonstrates exceptional performance in limited FOV scenarios, surpassing CMT by 6.7\% in mAP and 4.3\% in NDS. These results highlight MoME's effectiveness in handling partial sensor failures, leveraging the local attention mask in AQR.

\begin{table}[!t]
  \centering
  \footnotesize 
  \begin{tabular}{@{}c@{\hskip 0.2cm}c@{\hskip 0.2cm} cc@{\hskip 0.2cm}cc@{}}
    \toprule
    \multirow{2}[2]{*}{Method} & \multirow{2}[2]{*}{Modality} & \multicolumn{2}{c}{Rainy} & \multicolumn{2}{c}{Night} \\
    \cmidrule(lr){3-4} \cmidrule(lr){5-6}
    & & \multicolumn{1}{c}{mAP} & \multicolumn{1}{c}{NDS} & \multicolumn{1}{c}{mAP} & \multicolumn{1}{c}{NDS} \\
    \midrule
    BEVDet \cite{bevdet} & C & 30.1 & 43.2 & 12.0 & 21.7 \\
    BEVFormer \cite{bevformer} & C & 44.0 & - & 21.2 & - \\
    PETR \cite{petr} & C & 41.9 & 50.6 & 17.2 & 24.2 \\
    CenterPoint \cite{centerpoint} & L & 59.2 & - & 35.4 & - \\
    Transfusion-L \cite{transfusion} & L & 64.0 & 69.9 & 37.5 & 43.5 \\
    Transfusion \cite{transfusion} & LC & 67.5 & 71.9 & 39.8 & 44.7 \\
    Deepinteraction \cite{deepinteraction} & LC & 69.4 & 70.6 & 42.3 & 43.8 \\
    BEVFusion \cite{bevfusion_MIT} & LC & 69.9 & - & 42.8 & - \\
    GraphBEV \cite{graphbev} & LC & 70.2 & - & \textbf{45.1} & - \\
    
    CMT \cite{cmt} & LC & 70.5 & 73.7 & 42.8 & 46.3 \\
    UniTR$^{\dagger}$ \cite{unitr} & LC &71.1  & 74.4 & 36.6 & 43.3 \\
    SparseFusion$^{\dagger}$ \cite{sparsefusion} & LC & 70.2 & 73.3 & 43.8 & 46.4 \\
    \textbf{MoME (ours)} & LC & \textbf{72.2} & \textbf{74.8} & 43.0 & \textbf{46.6} \\
    \bottomrule
  \end{tabular}
  \caption{\textbf{Comparison under adverse weather conditions on nuScenes validation set.} `L' and `C' represent camera, LiDAR, respectively. }
  \label{tab_3}
\end{table}

\begin{table}[!t]
  \centering
  \footnotesize 
  
  \begin{tabular}{@{}ccccc@{}}
    \toprule
    \multicolumn{1}{c}{Method} & \multicolumn{1}{c}{Modality} & Fog & Snow & Sunlight \\
    \midrule
    PGD \cite{PGD} & C & 12.8 & 2.3 & 22.8 \\
    FCOS3D \cite{fcos3d} & C & 13.5 & 2.0 & 17.2 \\
    DETR3D \cite{detr3d} & C & 27.9 & 5.1 & 41.6 \\
    BEVFormer \cite{bevformer} & C & 32.8 & 5.7 & 41.7 \\
    PointPillars \cite{pointpillar} & L & 24.5 & 27.6 & 23.7 \\
    SSN \cite{ssn} & L & 41.6 & 46.4 & 40.3 \\
    CenterPoint \cite{centerpoint} & L & 43.8 & 55.9 & 54.2 \\
    FUTR3D \cite{futr3d} & LC & 53.2 & 52.7 & 57.7 \\
    Transfusion \cite{transfusion} & LC & 53.7 & 63.3 & 55.1 \\
    BEVFusion \cite{bevfusion_MIT} & LC & 54.1 & 62.8 & 64.4 \\
    DeepInteraction \cite{deepinteraction} & LC & 54.8 & 62.4 & 64.9 \\
    CMT \cite{cmt} & LC & 66.3 & 62.6 & 63.6 \\
    \textbf{MoME (ours)} & LC & \textbf{67.9} & \textbf{63.5} & \textbf{65.2} \\
    \bottomrule
  \end{tabular}
  \caption{\textbf{Comparison under extreme weather conditions on nuScenes-C.} }
  \label{tab_4}
\end{table}

Table \ref{tab_3} presents the performance of MoME under rainy and nighttime conditions on the nuScenes validation set. We used the same model as in Table \ref{table_2} without retraining. Despite not being explicitly trained for these scenarios, MoME demonstrated superior performance.

Under rainy conditions, where both LiDAR and camera sensors are significantly affected, MoME outperforms existing methods, achieving 72.2\% mAP and 74.8\% NDS. In nighttime conditions, MoME maintains competitive performance despite the challenges posed by low visibility. These results highlight MoME's adaptability to diverse weather conditions, with particularly strong robustness in rainy environments.

Table \ref{tab_4} presents MoME’s performance under extreme weather conditions on the nuScenes-C dataset. MoME consistently outperforms other methods, achieving mAP gains of +1.6\%, +0.9\%, and +1.6\% over CMT under fog, snow, and sunlight conditions, respectively.

\begin{table}[!t]
    \centering
    \footnotesize
    \label{ablation_1}
    \renewcommand{\arraystretch}{1.2}
     \begin{tabular}{@{}ccc|cc|cc|c@{}}
       \toprule
       \multicolumn{3}{c|}{Method} & \multicolumn{2}{c|}{Clean}& \multicolumn{2}{c|}{Perf. Ratio ($R$)} & \multicolumn{1}{c}{Latency }\\ 
        \cline{1-8}
        \multicolumn{1}{c}{ME} & \multicolumn{1}{c}{AQR}&\multicolumn{1}{c|}{LAM}& \multicolumn{1}{c}{mAP} & \multicolumn{1}{c|}{NDS}& \multicolumn{1}{c}{mAP} & \multicolumn{1}{c|}{NDS}& \multicolumn{1}{c}{ms}\\

        \hline
          & & & 70.3&72.9& 78.4&84.4 &\multicolumn{1}{c}{270} \\
          
          \checkmark & & &33.4&51.8& 113 & 99.5& \multicolumn{1}{c}{310} \\
        
         \checkmark& \checkmark & &71.0&73.6& 74.1 &83.2 & \multicolumn{1}{c}{274} \\

        \checkmark & \checkmark & \checkmark &\textbf{71.2}&\textbf{73.6}&  \textbf{80.5}& \textbf{86.2}&\multicolumn{1}{c}{274} \\
        
        \bottomrule
    \end{tabular}
    
    \caption{\textbf{Ablation studies on MoME components.} ME denotes Multiple Expert decoders, AQR denotes Adaptive Query Router, and LAM indicates Local Attention Mask. 
}
\label{ablation_1}
\end{table}

\begin{table}[!t]
    \centering
    \scriptsize
    \label{ablation_query}
    \renewcommand{\arraystretch}{1.2}
     \begin{tabular}{@{}c|ccc@{}}
       \toprule
       Allocated queries & LiDAR-Cam Expert&LiDAR Expert&Camera Expert\\ 
        \hline
        
        Clean& 94\% &0\% &6\%  \\
        LiDAR Drop& 8\% &0\% &92\%  \\
        Camera Drop& 0\% &100\% &0\%  \\
        Limited FOV& 38\% &0\% &62\%  \\
        Occlusion& 81\% &15\% &4\%  \\
        Object Failure& 90\% &0\% &10\%  \\
        Fog& 99\% &0\% &1\%  \\
        Snow& 95\% &0\% &5\%  \\
        Sunlight& 94\% &0\% &6\%  \\
        Rainy& 99\% &0\% &1\%  \\
        Night& 99\% &0\% &1\%  \\
        \bottomrule
    \end{tabular}
    \caption{\textbf{Query allocation results across different failure scenarios.} The table presents the percentage of query selections across different scenarios.}
\label{ablation_query_selection}
\end{table}

\begin{figure*}[h]
    \centering
    {\includegraphics[width=0.95\textwidth]{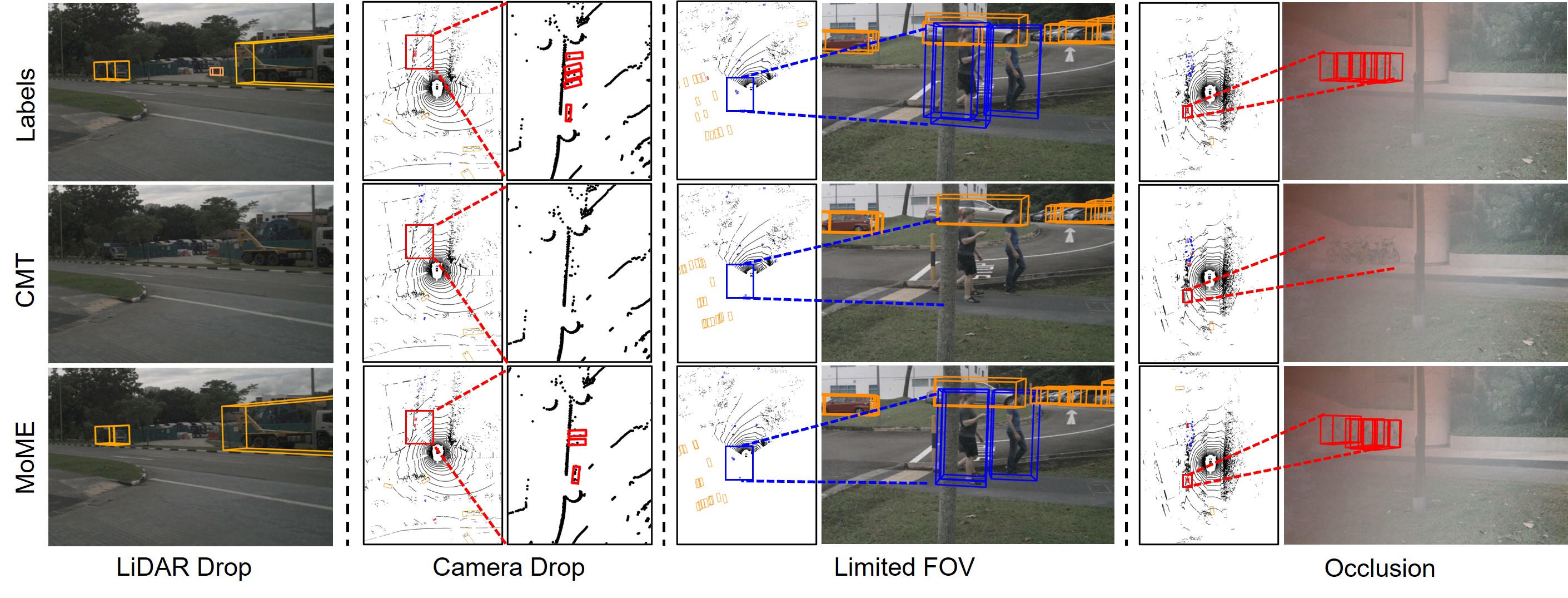}}
    \hspace{5mm}
    \caption {\textbf{Qualitative results under various sensor failure scenarios.}  Comparison of detection results under four challenging scenarios: LiDAR drop, Camera drop, Limited FOV, and Occlusion. MoME exhibits higher robustness and consistent results compared to CMT~\cite{cmt}}
    \label{fig2}
\end{figure*} 

\subsection{Ablation Studies}

\noindent\textbf{Contribution of Main Components.} In this section, we analyze the contribution of each component to overall performance. Table \ref{ablation_1} presents an ablation study conducted on the nuScenes validation set for the clean case and the nuScenes-R set for failure cases.

We use the single decoder as our baseline, adopting the architecture of CMT \cite{cmt}. When incorporating Multiple Experts (ME), we utilize separate decoders without a routing procedure. Among the outputs from these decoders, we select one based on its confidence score. However, this approach leads to a significant performance drop due to uncalibrated confidence scores from independently optimized decoders. Although the relative performance ratio $R$ appears to improve, this is primarily due to a considerable decline in performance for the clean case, rather than an actual increase in robustness.

Next, we introduce the AQR module, applying global attention while omitting the local attention mechanism of LAM. AQR effectively resolves the score calibration issue and restores performance by leveraging adaptive query routing. However, without LAM, the performance ratio $R$ does not surpass that of the baseline, suggesting that decoder selection is still suboptimal.

Finally, enabling both AQR and LAM leads to a substantial improvement in the performance ratio $R$, resulting in an increase of 6.4 in mAP and 3.0 in NDS. Consequently, our complete MoME model achieves an overall improvement of 2.1 in mAP and 1.8 in NDS compared to the baseline method.

Notably, our proposed method increases latency by only $4$ ms over the baseline, which is lower than $40$ ms of the parallel decoder with ME. This efficiency is attributed to the distributed query processing mechanism of MoME.

\noindent\textbf{Query Allocation Behavior of AQR.} 
Table \ref{ablation_query_selection} presents the statistical distribution of decoder selection across various sensor failure scenarios. For each case, we count the number of queries assigned to each decoder by the AQR module. Under clean conditions, our AQR module predominantly routes 94\% of queries to the LiDAR-Camera expert, maximizing the benefits of sensor fusion. Notably, even in clean scenarios, 6\% of queries are allocated to the Camera expert, likely due to sparse LiDAR points for small or distant objects.

In LiDAR Drop scenarios, the AQR module redirects 92\% of queries to the Camera expert, effectively avoiding the degraded modality. In Camera Drop scenarios, AQR allocates 100\% of queries to the LiDAR expert. For partial sensor degradation, such as Limited FOV, AQR assigns 38\% of queries to the LiDAR-Camera decoder and 62\% to the Camera decoder. In Occlusion scenarios, the model maintains robust performance by allocating 81\% of queries to the LiDAR-Camera decoder and 15\% to the LiDAR decoder.
Environmental challenges such as fog, snow, and sunlight result in allocation patterns similar to those in clean conditions. However, instances where the Camera decoder is selected may play a critical role in ensuring robust performance.

\noindent\textbf{Qualitative Results.} 
Fig. \ref{fig2} presents qualitative results comparing our proposed MoME architecture with the SOTA method, CMT \cite{cmt} across various sensor failure scenarios. While CMT is significantly impacted by sensor failures, leading to inaccurate detection results, MoME achieves more robust detection by leveraging multiple expert decoders, demonstrating the advantages of architectural independence.
\section{Conclusions}

In this paper, we introduced \textit{MoME}, an efficient multi-modal fusion framework designed to enhance robustness under sensor failure scenarios. Our approach adopts the Mixture of Experts concept to exploit redundancy in decoding camera and LiDAR features. Specifically, we employed three decoders—Camera, LiDAR, and Camera-LiDAR—that process camera features, LiDAR features, and both modalities, respectively. Our MoME framework selects the appropriate decoder for each query using AQR, which performs locality-aware routing based on local features identified by a local attention mask.

To ensure robust performance, we trained the MoME model with random sensor dropouts, allowing it to adapt to missing camera or LiDAR inputs. Our evaluation on the nuScenes benchmark demonstrated that MoME achieves strong robustness across various sensor failure scenarios and adverse environmental conditions in the nuScenes-R and nuScenes-C benchmarks. Notably, our method established a new state-of-the-art performance.

For future work, we aim to develop more effective training strategies to simulate diverse failure patterns, further enhancing robustness against rarely occurring sensor failures.

\section{Acknowledgements}
This work was partly supported by 1) Institute of Information \& communications Technology Planning \& Evaluation (IITP) grant funded by the Korea government(MSIT) [NO.RS-2021-II211343, Artificial Intelligence Graduate School Program (Seoul National University)] and 2) the National Research Foundation (NRF) funded by the Korean government (MSIT) (No. RS-2024-00421129).
{
    \small
    \bibliographystyle{ieeenat_fullname}
    \bibliography{main}
}
\clearpage
\setcounter{page}{1}
\maketitlesupplementary
\setcounter{section}{0}  
\renewcommand{\thesection}{\Alph{section}}
We present detailed pseudo-code to describe the MoME framework and AQR mechanism in MoME. In the following sections, we detail the implementation specifications of MoME. The performance of MoME is thoroughly evaluated across various sensor failure scenarios. Our comprehensive qualitative analysis further validates MoME's robustness and performance advantages.

\section{Algorithms}
We provide detailed pseudo-code with PyTorch including AQR and RAM in Algorithm~\ref{alg:aqr} and Algorithm~\ref{alg:ram}.

\section{Implementation Details}
\subsection{Training Details}
The training strategy of MoME consists of two stages to handle sensor failures. In the first stage, all object queries in $Q$ are processed in parallel by each expert decoder and matched with ground truth through bipartite matching, without any sensor drop augmentation. The second stage focuses on handling sensor failures by applying sensor drop augmentation, where we randomly mask either the camera or LiDAR inputs, with each sensor having a $1/3$ probability of being dropped, while retaining both sensors for the remaining $1/3$ of cases.
\subsection{Adverse Sensor Scenarios}
Our experimental validation encompasses both sensor failure scenarios and adverse weather conditions. Our sensor failure experiments incorporate BEVFusion's~\cite{bevfusion_MIT} Beam Reduction settings and nuScenes-R's~\cite{robustbench} protocols for LiDAR Drop, Limited FOV, Object Failure, View Drop, and Occlusion. For adverse weather conditions, we utilize scene descriptions in the nuScenes~\cite{nuscenes} validation set to identify Rainy and Night scenarios, while adopting Fog, Snow, and Sunlight conditions from nuScenes-C~\cite{benchmarking_cc}.
 
\begin{table*}[!h]
    \centering
    \tiny
    \renewcommand{\arraystretch}{1.15}
    \setlength{\arrayrulewidth}{0.3pt}
    \resizebox{0.9\textwidth}{!}{%
     \begin{tabular}{@{}c|cc|cc|cc|cc|cc|cc@{\hspace{4pt}}}
       \toprule[0.1pt]
       \multirow{2}{*}[-2pt]{\tiny \hspace{2pt} Method} & \multicolumn{2}{c|}{Limited FOV} & \multicolumn{2}{c|}{Limited FOV} & \multicolumn{2}{c|}{Limited FOV} & \multicolumn{2}{c|}{Limited FOV} & \multicolumn{2}{c|}{Limited FOV} & \multicolumn{2}{c}{Limited FOV} \\ 
        
        \multicolumn{1}{c|}{} & \multicolumn{2}{c|}{\textit{[-180, 180]}} & \multicolumn{2}{c|}{\textit{[-150, 150]}} & \multicolumn{2}{c|}{\textit{[-120, 120]}} & \multicolumn{2}{c|}{\textit{[-90, 90]}} & \multicolumn{2}{c|}{\textit{[-60, 60]}} & \multicolumn{2}{c}{\textit{[-30, 30]}} \\
        \cline{2-13}
        & mAP & NDS & mAP & NDS & mAP & NDS & mAP & NDS & mAP & NDS & mAP & NDS \\ 
        \hline
        \hspace{2pt} CMT~\cite{cmt} & 70.3 & 72.9 & 65.4 & 69.8 & 53.8 & 62.2 & 49.0 & 58.4 & 43.9 & 54.0 & 35.0 & 46.1 \\
        \textbf{ \hspace{2pt} MoME (ours)} & \textbf{71.2} & \textbf{73.6} & \textbf{67.8} & \textbf{71.2} & \textbf{56.2} & \textbf{63.4} & \textbf{54.2} & \textbf{61.2} & \textbf{50.6} & \textbf{58.3} & \textbf{44.0} & \textbf{53.0} \\ 
        \bottomrule[0.01pt]
    \end{tabular}
    }
    \caption{Performance comparison between CMT \cite{cmt} and MoME (ours) on \textit{Limited FOV}}
\label{table_1_s}
\end{table*}
\begin{table*}[!h]
    \centering
    \tiny
    \renewcommand{\arraystretch}{1.15}
    \setlength{\arrayrulewidth}{0.3pt}
    \resizebox{0.9\textwidth}{!}{%
     \begin{tabular}{@{}c|cc|cc|cc|cc|cc|cc@{\hspace{4pt}}}
       \toprule[0.1pt]
       \multirow{2}{*}[-2pt]{\tiny \hspace{2pt} Method} & \multicolumn{2}{c|}{Object Failure} & \multicolumn{2}{c|}{Object Failure} & \multicolumn{2}{c|}{Object Failure} & \multicolumn{2}{c|}{Object Failure} & \multicolumn{2}{c|}{Object Failure} & \multicolumn{2}{c}{Object Failure} \\ 
        
        \multicolumn{1}{c|}{} & \multicolumn{2}{c|}{\textit{ratio=0.0}} & \multicolumn{2}{c|}{\textit{ratio=0.1}} & \multicolumn{2}{c|}{\textit{ratio=0.3}} & \multicolumn{2}{c|}{\textit{ratio=0.5}} & \multicolumn{2}{c|}{\textit{ratio=0.7}} & \multicolumn{2}{c}{\textit{ratio=0.9}} \\
        \cline{2-13}
        & mAP & NDS & mAP & NDS & mAP & NDS & mAP & NDS & mAP & NDS & mAP & NDS \\ 
        \hline
        \hspace{2pt} CMT~\cite{cmt} & 70.3 & 72.9 & 68.8 & 71.6 & 67.6 & 70.8 & 66.7 & 70.4 & 64.5 & 68.2 & 62.7 & 67.2 \\
        \textbf{ \hspace{2pt} MoME (ours)} & \textbf{71.2} & \textbf{73.6} & \textbf{70.0} & \textbf{72.8} & \textbf{68.5} & \textbf{71.8} & \textbf{67.0} & \textbf{71.0} & \textbf{64.8} & \textbf{68.8} & \textbf{63.0} & \textbf{67.8} \\ 
        \bottomrule[0.01pt]
    \end{tabular}
    }
    \caption{Performance comparison between CMT \cite{cmt} and MoME (ours) on \textit{Object Failure}}
\label{table_2_s}
\end{table*}
\begin{table*}[!h]
    \centering
    \tiny
    \renewcommand{\arraystretch}{1.15}
    \setlength{\arrayrulewidth}{0.3pt}
    \resizebox{0.9\textwidth}{!}{%
     \begin{tabular}{@{}c|cc|cc|cc|cc|cc@{\hspace{4pt}}}
       \toprule[0.1pt]
       \multirow{2}{*}[-2pt]{\tiny \hspace{2pt} Method} & \multicolumn{2}{c|}{Beam Reduction} & \multicolumn{2}{c|}{Beam Reduction} & \multicolumn{2}{c|}{Beam Reduction} & \multicolumn{2}{c|}{Beam Reduction} & \multicolumn{2}{c}{Beam Reduction} \\ 
        
        \multicolumn{1}{c|}{} & \multicolumn{2}{c|}{\textit{1 beams}} & \multicolumn{2}{c|}{\textit{4 beams}} & \multicolumn{2}{c|}{\textit{8 beams}} & \multicolumn{2}{c|}{\textit{16 beams}} & \multicolumn{2}{c}{\textit{32 beams}} \\
        \cline{2-11}
        & mAP & NDS & mAP & NDS & mAP & NDS & mAP & NDS & mAP & NDS \\ 
        \hline
        \hspace{2pt} CMT~\cite{cmt} & 25.9 & 42.6 & 54.9 & 62.2 & 59.5 & 65.4 & 62.3 & 67.5 & 70.3 & 72.9 \\
        \textbf{ \hspace{2pt} MoME (ours)} & \textbf{30.5} & \textbf{43.4} & \textbf{55.0} & \textbf{63.0} & \textbf{60} & \textbf{66.5} & \textbf{62.7} & \textbf{68.3} & \textbf{71.2} & \textbf{73.6} \\ 
        \bottomrule[0.01pt]
    \end{tabular}
    }
    \caption{Performance comparison between CMT \cite{cmt} and MoME (ours) on \textit{Beam Reduction}}
\label{table_3_s}
\end{table*}
\begin{table*}[!h]
    \centering
    \tiny
    \renewcommand{\arraystretch}{1.15}
    \setlength{\arrayrulewidth}{0.3pt}
    \resizebox{0.9\textwidth}{!}{%
     \begin{tabular}{@{}c|cc|cc|cc|cc|cc|cc@{\hspace{4pt}}}
       \toprule[0.1pt]
       \multirow{2}{*}[-2pt]{\tiny \hspace{2pt} Method} & \multicolumn{2}{c|}{View Drop} & \multicolumn{2}{c|}{View Drop} & \multicolumn{2}{c|}{View Drop} & \multicolumn{2}{c|}{View Drop} & \multicolumn{2}{c|}{View Drop} & \multicolumn{2}{c}{View Drop} \\ 
        
        \multicolumn{1}{c|}{} & \multicolumn{2}{c|}{\textit{1 drop}} & \multicolumn{2}{c|}{\textit{2 drops}} & \multicolumn{2}{c|}{\textit{3 drops}} & \multicolumn{2}{c|}{\textit{4 drops}} & \multicolumn{2}{c|}{\textit{5 drops}} & \multicolumn{2}{c}{\textit{6 drops}} \\
        \cline{2-13}
        & mAP & NDS & mAP & NDS & mAP & NDS & mAP & NDS & mAP & NDS & mAP & NDS \\ 
        \hline
        \hspace{2pt} CMT~\cite{cmt} & 68.1 & 71.7 & 67.1 & 71.1 & 65.6 & 70.4 & 64.0 & 69.5 & 62.6 & 68.6 & 61.7 & 68.1 \\
        \textbf{ \hspace{2pt} MoME (ours)} & \textbf{68.6} & \textbf{72.4} & \textbf{67.7} & \textbf{71.9} & \textbf{66.6} & \textbf{71.2} & \textbf{64.6} & \textbf{70.4} & \textbf{63.8} & \textbf{69.9} & \textbf{63.6} & \textbf{69.5} \\ 
        \bottomrule[0.01pt]
    \end{tabular}
    }
    \caption{Performance comparison between CMT \cite{cmt} and MoME (ours) on \textit{View Drop}}
\label{table_4_s}
\end{table*}

\section{Extensive Performance Comparisons}

We present additional experimental results by extending our analysis across different parameter settings for each sensor failure scenario. While Table~\ref{table_2} shows the results with fixed configurations, Tables~\ref{table_1_s}-\ref{table_4_s} provide comprehensive evaluations with various parameter ranges for each failure case.
\begin{itemize}
    \item \textbf{Limited FOV} (Table~\ref{table_1_s}): We observe that MoME demonstrates better performance gains over CMT~\cite{cmt} as the field of view becomes more restricted. While both methods achieve comparable mAP scores of 71.2\% and 70.3\% respectively in the full FOV range of \textit{[-180, 180]}, the performance gap widens significantly under severe FOV limitations, where MoME achieves 44.0\% mAP compared to CMT's 35.0\% mAP at \textit{[-30, 30]}, demonstrating MoME's superior robustness to limited FOV.
    \item \textbf{Object Failure} (Table~\ref{table_2_s}): We evaluate MoME and CMT with different object failure ratios. Specifically, our method achieves 72.8\% NDS, outperforming CMT which achieves 71.6\% NDS at \textit{ratio=0.1}.
    \item \textbf{Beam Reduction} (Table~\ref{table_3_s}): We analyze performance from \textit{1} to \textit{32 beams}, showing significant improvements especially with reduced beams, as our method achieves 30.5\% mAP while CMT reaches 25.9\% mAP using \textit{1 beam}.
    \item \textbf{View Drop} (Table~\ref{table_4_s}): MoME and CMT show gradual performance degradation as the number of dropped views increases. MoME maintains consistently higher performance, achieving 68.6\% mAP compared to CMT's 68.1\% mAP with \textit{1 drop}, and the performance gap widens with \textit{6 drops}, where MoME achieves 63.6\% mAP while CMT reaches 61.7\% mAP.
\end{itemize}
\begin{table*}[!h]
    \centering
    \footnotesize
    \label{ablation_3}
    \renewcommand{\arraystretch}{1.15}
    \resizebox{\textwidth}{!}{%
     \begin{tabular}{@{}c|cc|cc|cc|cc|cc|cc|cc@{}}
       \toprule
       \multicolumn{1}{c|}{} & \multicolumn{2}{c|}{}&\multicolumn{8}{c|}{LiDAR failure} & \multicolumn{4}{c}{Camera failure}\\
       \cline{4-11} \cline{12-15}
       \multirow{2}[-1]{*}{Method} & \multicolumn{2}{c|}{Clean} & \multicolumn{2}{c|}{Beam Reduction} & \multicolumn{2}{c|}{LiDAR Drop} & \multicolumn{2}{c|}{Limited FOV} & \multicolumn{2}{c|}{Object Failure} & \multicolumn{2}{c|}{View Drop} & \multicolumn{2}{c}{Occlusion} \\ 
        
         & \multicolumn{2}{c|}{} & \multicolumn{2}{c|}{\textit{4 beams}} & \multicolumn{2}{c|}{\textit{all }} & \multicolumn{2}{c|}{\textit{[-60, 60]}} & \multicolumn{2}{c|}{\textit{rate = 0.5}} & \multicolumn{2}{c|}{\textit{6 drops}}& \multicolumn{2}{c}{\textit{w obstacle}} \\
        \cline{2-15}
        & \multicolumn{1}{c}{mAP} & \multicolumn{1}{c|}{NDS} & \multicolumn{1}{c}{mAP} & \multicolumn{1}{c|}{NDS} & \multicolumn{1}{c}{mAP} & \multicolumn{1}{c|}{NDS} & \multicolumn{1}{c}{mAP} & \multicolumn{1}{c|}{NDS} & \multicolumn{1}{c}{mAP} & \multicolumn{1}{c|}{NDS} & \multicolumn{1}{c}{mAP} & \multicolumn{1}{c|}{NDS} & \multicolumn{1}{c}{mAP} & \multicolumn{1}{c}{NDS} \\

        \hline
        
        cross attention & \multicolumn{1}{c}{71.0} & \multicolumn{1}{c|}{73.6} & \multicolumn{1}{c}{54.6} & \multicolumn{1}{c|}{62.8} & \multicolumn{1}{c}{42.3} & \multicolumn{1}{c|}{48.1} & \multicolumn{1}{c}{25.7} & \multicolumn{1}{c|}{48.0} & \multicolumn{1}{c}{66.0} & \multicolumn{1}{c|}{69.8} & \multicolumn{1}{c}{63.0} & \multicolumn{1}{c|}{69.4} & \multicolumn{1}{c}{64.1} & \multicolumn{1}{c}{69.3} \\

        deformable attention& \multicolumn{1}{c}{71.0} & \multicolumn{1}{c|}{73.6} & \multicolumn{1}{c}{54.8} & \multicolumn{1}{c|}{62.9} & \multicolumn{1}{c}{42.3} & \multicolumn{1}{c|}{48.1} & \multicolumn{1}{c}{32.1} & \multicolumn{1}{c|}{50.1} & \multicolumn{1}{c}{66.0} & \multicolumn{1}{c|}{70.1} & \multicolumn{1}{c}{63.1} & \multicolumn{1}{c|}{69.4} & \multicolumn{1}{c}{64.0} & \multicolumn{1}{c}{69.6} \\

        MLP & \multicolumn{1}{c}{70.7} & \multicolumn{1}{c|}{73.5} & \multicolumn{1}{c}{54.8} & \multicolumn{1}{c|}{62.9} & \multicolumn{1}{c}{42.3} & \multicolumn{1}{c|}{48.1} & \multicolumn{1}{c}{44.2} & \multicolumn{1}{c|}{54.4} & \multicolumn{1}{c}{66.4} & \multicolumn{1}{c|}{70.3} & \multicolumn{1}{c}{63.1} & \multicolumn{1}{c|}{69.5} & \multicolumn{1}{c}{64.6} & \multicolumn{1}{c}{69.8} \\

        AQR (ours) & \multicolumn{1}{c}{\textbf{71.2}} & \multicolumn{1}{c|}{\textbf{73.6}} & \multicolumn{1}{c}{\textbf{55.0}} & \multicolumn{1}{c|}{\textbf{63.0}} & \multicolumn{1}{c}{\textbf{42.5}} & \multicolumn{1}{c|}{\textbf{48.2}} & \multicolumn{1}{c}{\textbf{50.6}} & \multicolumn{1}{c|}{\textbf{58.3}} & \multicolumn{1}{c}{\textbf{67.0}} & \multicolumn{1}{c|}{\textbf{71.0}} & \multicolumn{1}{c}{\textbf{63.6}} & \multicolumn{1}{c|}{\textbf{69.5}} & \multicolumn{1}{c}{\textbf{65.6}} & \multicolumn{1}{c}{\textbf{70.5}} \\
        
        \bottomrule
    \end{tabular}
    }
    \caption{\textbf{Ablation Studies on feature extraction methods for AQR query routing.} AQR with LAM shows the superior robustness, particularly under local-aware sensor failure scenarios, such as Limited FOV, Object Failure, and Occlusion.}
\label{ablation_3}
\end{table*}
\section{Local Feature Extraction Methods.}  
 The AQR module selects one of the expert decoders based on features extracted from a local region identified by a query. In Table \ref{ablation_3}, we compare different feature extraction methods within our router architecture, including cross-attention, deformable attention, MLP, and our proposed approach. 
 These methods exhibit notable performance differences in Limited FOV scenarios, where partial sensor failure occurs. This is because selecting an appropriate decoder depends on the query's position. Notably, cross-attention demonstrates limited effectiveness, as it struggles to focus specifically on locally degraded regions. Deformable attention performs better due to its dynamic spatial sampling capability, but it lacks explicit control over attention regions. While the MLP approach, which utilizes max-pooled multi-modal features, shows reasonable robustness to failure cases, it underperforms in the Clean scenario. In contrast, our proposed AQR module consistently achieves superior performance across all cases, thanks to its use of a local attention mask.

\section{Additional Qualitative Results}
\begin{figure*}[h]
    \centering
    {\includegraphics[width=0.95\textwidth]{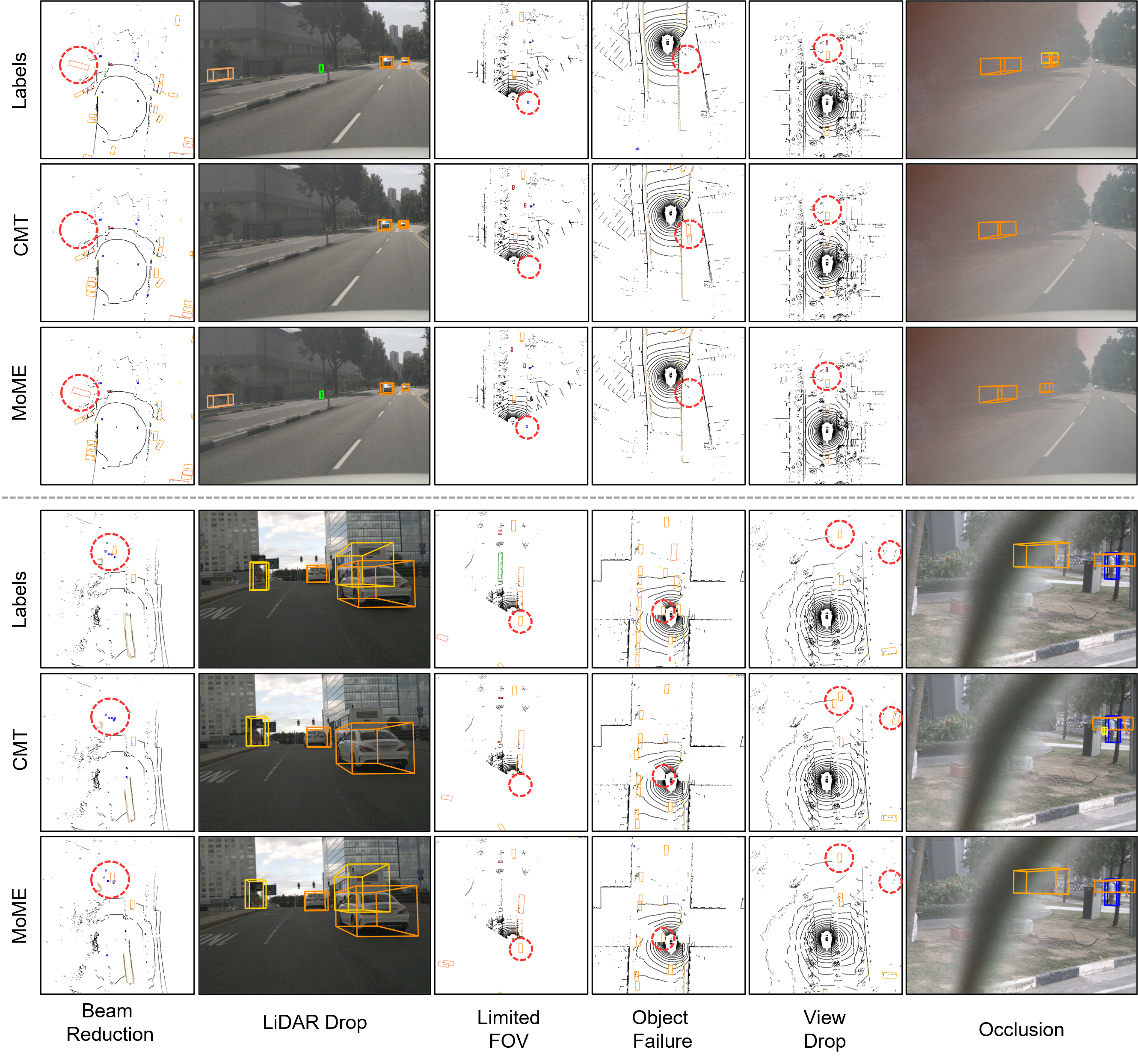}}
    \hspace{5mm}
    \caption {\textbf{Qualitative results under various sensor failure scenarios.}  Comparison of detection results between MoME and CMT~\cite{cmt} under six sensor failure scenarios: Beam Reduction, LiDAR Drop, Limited FOV, Object Failure, View Drop, and Occlusion. The results demonstrate MoME's detection capabilities across these challenging conditions.}
    \label{fig3}
\end{figure*}

Fig.~\ref{fig3} compares the detection results of MoME and CMT~\cite{cmt} under six sensor failure scenarios in real-world scenes. The visualizations demonstrate our model's consistent performance across different failure conditions.
\begin{itemize}
\item \textbf{Beam Reduction}: Even with reduced LiDAR input, MoME accurately detects objects in the first three rows and successfully captures vehicles behind pedestrians in the last three rows.
\item \textbf{LiDAR Drop}: MoME effectively leverages camera information to detect small and partially occluded objects that CMT fails to identify.
\item \textbf{Limited FOV}: MoME successfully detects small objects in LiDAR-absent regions compared to CMT.
\item \textbf{Object Failure}: MoME achieves lower false positive rates than CMT and maintains accurate detection of nearby objects under object failure conditions.
\item \textbf{View Drop}: MoME successfully detects occluded small objects that CMT misses and reduces false positive detections in complex scenes with dropped camera views.
\item \textbf{Occlusion}:When parts of the scene are masked to simulate occlusions, MoME successfully detects objects in the affected regions while CMT shows degraded performance.
\end{itemize}

\definecolor{pinkish}{RGB}{255,192,203}
\definecolor{bluish}{RGB}{70,130,180}
\definecolor{orangeish}{RGB}{255,140,0}

\lstdefinestyle{customcode}{
    basicstyle=\ttfamily\small,
    breaklines=false,
    breakatwhitespace=false,
    frame=none,              
    numbers=none,
    showstringspaces=false,
    columns=fixed,
    basewidth=0.5em,
    keepspaces=true,
    tabsize=4,
    commentstyle=\color{gray},
    keywordstyle=\color{black},
    stringstyle=\color{black},
    backgroundcolor=\color{white},
    xleftmargin=0pt,
    xrightmargin=0pt,
    linewidth=\textwidth,
    basicstyle=\ttfamily\small\color{black},
    morekeywords=[1]{import, class, def, super, if, elif, else, return, from},
    keywordstyle=[1]{\color{pinkish}},
    morekeywords=[2]{and, not},
    keywordstyle=[2]{\color{bluish}},
    keywordstyle=[3]{\color{orangeish}},
    literate={\#}{{{\color{bluish}\#}}}1,
}

\begin{figure*}[t]
\begin{algorithm}[H]
\caption{Adaptive Query Router (AQR)}
\begin{lstlisting}[style=customcode]
import torch
import torch.nn as nn
from mmcv.cnn.bricks.transformer import build_transformer_layer_sequence

class AQR(nn.Module):
    def __init__(self, encoder, hidden_dim, **kwargs):
        super().__init__()
        self.encoder = build_transformer_layer_sequence(encoder)
        self.linear = nn.Linear(hidden_dim, 3)
       
    def forward(self, c_dict, ref_pts, pc_range, img_feats, metas):
        # 3D -> 2D projection.
        rp = torch.stack([ref_pts[..., i:i+1] * (pc_range[i+3] - pc_range[i]) + pc_range[i] 
                       for i in range(3)], -1)
        b, n = rp.shape[:2]
        p2d = torch.einsum('bni,bvij->bvnj', 
                      torch.cat([rp, torch.ones((b, n, 1)).cuda()], -1),
                      torch.tensor(np.stack([np.stack(i['lidar2img']) for i in metas]))
                      .float().cuda().transpose(2,3))
        p2d[..., :2] /= torch.clip(p2d[..., 2:3], min=1e-5)

        # Get valid points & prepare inputs.
        h, w = metas[0]['img_shape'][0][:2]
        v = ((p2d[..., 0] < w) & (p2d[..., 0] >= 0) & (p2d[..., 1] < h) & (p2d[..., 1] >= 0))
        v = (torch.cat([torch.zeros_like(v[:,:1,:], dtype=torch.bool), v], 1)
           .float().argmax(1) - 1) * (v != 0)
      
        m = v != -1
        p_cam = torch.zeros((b, n, 3), device=p2d.device)
        p_cam[m] = torch.cat([v[m].unsqueeze(-1), 
                p2d[torch.where(m)[0], v[m], torch.where(m)[1]][...,[1,0]] * 
                (img_feats.shape[2] / h)], -1)
        p_lidar = torch.floor((rp[..., :2] + 54.0) * (180 / 108))[..., [1,0]] 
        attention_mask = [RAM(p_lidar, p_cam, b, n).unsqueeze(1)
                        .repeat(1, self.e_num_heads, 1, 1).flatten(0, 1)]
        # Forward.
        out = self.encoder(
            torch.zeros_like(c_dict['query_embed_l'][0]),
            c_dict['memory_l'][0],
            c_dict['memory_v_l'][0],
            c_dict['query_embed_l'][0],
            c_dict['pos_embed_l'][0],
            attention_mask)
        out = self.linear((out[-1] if out.shape[0] != 0 else out.squeeze(0))
                             .transpose(1, 0))
        celoss = nn.CrossEntropyLoss()
        loss = celoss(out.reshape(-1, 3), torch.tensor([i['modalmask'] for i in metas]).cuda()
        .unsqueeze(1).repeat(1, n, 1).reshape(-1, 3).float()) if self.training else None
        return out, loss
            
\end{lstlisting}
\label{alg:aqr}
\end{algorithm}
\end{figure*}

\begin{figure*}[t]
\begin{algorithm}[H]
\caption{Local Attention Mask}
\begin{lstlisting}[style=customcode]
import torch

def LAM(c_dict, p_lidar, p_cam, b, n):
    # LiDAR mask.
    rs, wl = 180, 5  # row stride, window size for lidar.
    l_idx = p_lidar[..., 0] * rs + p_lidar[..., 1]
    off = torch.arange(-(wl // 2), wl // 2 + 1, device=p_lidar.device)
    y, x = torch.meshgrid(off, off)
    l_win = (y * rs + x).reshape(-1)
    l_indices = l_idx.unsqueeze(-1) + l_win
    
    l_valid = (l_indices >= 0) & (l_indices < rs * rs) & \
             ((l_indices % rs - l_idx.unsqueeze(-1) % rs).abs() <= wl // 2)
    l_mask = torch.ones(b, n, rs * rs, dtype=torch.bool, device=p_lidar.device)
    
    # Camera mask.
    h, w, wc = 40, 100, 15  # height, width, window size for camera.
    c_idx = p_cam[..., 0] * h * w + p_cam[..., 1] * w + p_cam[..., 2]
    off = torch.arange(-(wc // 2), wc // 2 + 1, device=p_cam.device)
    c_win = (off.unsqueeze(1) * w + off.unsqueeze(0)).reshape(-1)
    c_indices = c_idx.unsqueeze(-1) + c_win
    
    qp = c_idx % (h * w)
    c_valid = ((c_indices % (h * w) // w - qp.unsqueeze(-1) // w).abs() <= wc // 2) & \
             ((c_indices % w - qp.unsqueeze(-1) % w).abs() <= wc // 2)
    c_mask = torch.ones(b, n, 6 * h * w, dtype=torch.bool, device=p_cam.device)
    c_indices = torch.clamp(c_indices, 0, 6 * h * w - 1)
    
    # Update masks with valid indices.
    bid = torch.arange(b, device=p_lidar.device).view(-1,1,1)
    qid = torch.arange(n, device=p_lidar.device).view(1,-1,1)
    l_mask[bid.expand_as(l_indices)[l_valid], qid.expand_as(l_indices)[l_valid], l_indices[l_valid]] = False
    c_mask[bid.expand_as(c_indices)[c_valid], qid.expand_as(c_indices)[c_valid], c_indices[c_valid]] = False
    
    return torch.cat([l_mask, c_mask], -1)
            
\end{lstlisting}
\label{alg:ram}
\end{algorithm}
\end{figure*}

\end{document}